\documentclass[letterpaper, 10 pt, conference]{ieeeconf}  

\IEEEoverridecommandlockouts                              

\makeatletter
\let\NAT@parse\undefined
\makeatother

\usepackage{amsmath}
\usepackage{graphicx,url}
\usepackage{caption}
\usepackage{subcaption}
\usepackage{cite}
\usepackage{balance}

\usepackage{hyperref}
\hypersetup{bookmarksopen,bookmarksnumbered,
pdfpagemode=UseOutlines,
colorlinks=true,
linkcolor=blue,
anchorcolor=blue,
citecolor=blue,
filecolor=blue,
menucolor=blue,
urlcolor=blue,
breaklinks=true
}

\usepackage{booktabs}

\title{\LARGE \bf
Optical Proximity Sensing for Pose Estimation\\ During In-Hand Manipulation
}

\author{Patrick Lancaster$^{1}$, Pratik Gyawali$^{2}$,  Christoforos Mavrogiannis$^{1}$, Siddhartha S. Srinivasa$^{1}$, Joshua R. Smith$^{1,3}$
\thanks{$^{1}$ Patrick Lancaster, Christoforos Mavrogiannis, Siddhartha S. Srinivasa, and Joshua R. Smith are with the Paul G. Allen School of Computer Science and Engineering,
        University of Washington, Seattle, WA 98195, USA
        {\tt\small planc509@cs.uw.edu, 
        cmavro@cs.uw.edu, 
        siddh@cs.uw.edu, 
        jrs@cs.uw.edu}}%
\thanks{$^{2}$ Pratik Gyawali is with the the Department of Mechanical Engineering, University of Washington,        Seattle, WA 98195, USA {\tt\small gyawali@uw.edu}}
\thanks{$^{3}$ Joshua R. Smith is with the Department of Electrical and Computer Engineering, University of Washington,
        Seattle, WA 98195, USA}%
}

\begin{document}

\maketitle
\thispagestyle{empty}
\pagestyle{empty}

\begin{abstract}

During in-hand manipulation, robots must be able to continuously estimate the pose of the object in order to generate appropriate control actions. The performance of algorithms for pose estimation hinges on the robot's sensors being able to detect discriminative geometric object features, but previous sensing modalities are unable to make such measurements robustly. The robot's fingers can occlude the view of environment- or robot-mounted image sensors, and tactile sensors can only measure at the local areas of contact. Motivated by fingertip-embedded proximity sensors' robustness to occlusion and ability to measure beyond the local areas of contact, we present the first evaluation of proximity sensor based pose estimation for in-hand manipulation. We develop a novel two-fingered hand with fingertip-embedded optical time-of-flight proximity sensors as a testbed for pose estimation during planar in-hand manipulation. Here, the in-hand manipulation task consists of the robot moving a cylindrical object from one end of its workspace to the other. We demonstrate, with statistical significance, that proximity-sensor based pose estimation via particle filtering during in-hand manipulation: a) exhibits $50\%$ lower average pose error than a tactile-sensor based baseline; b) empowers a model predictive controller to achieve $30\%$ lower final positioning error compared to when using tactile-sensor based pose estimates. 





\end{abstract}

\section{INTRODUCTION}

\begin{figure}[t]
\centering
\includegraphics[width=0.48\textwidth]{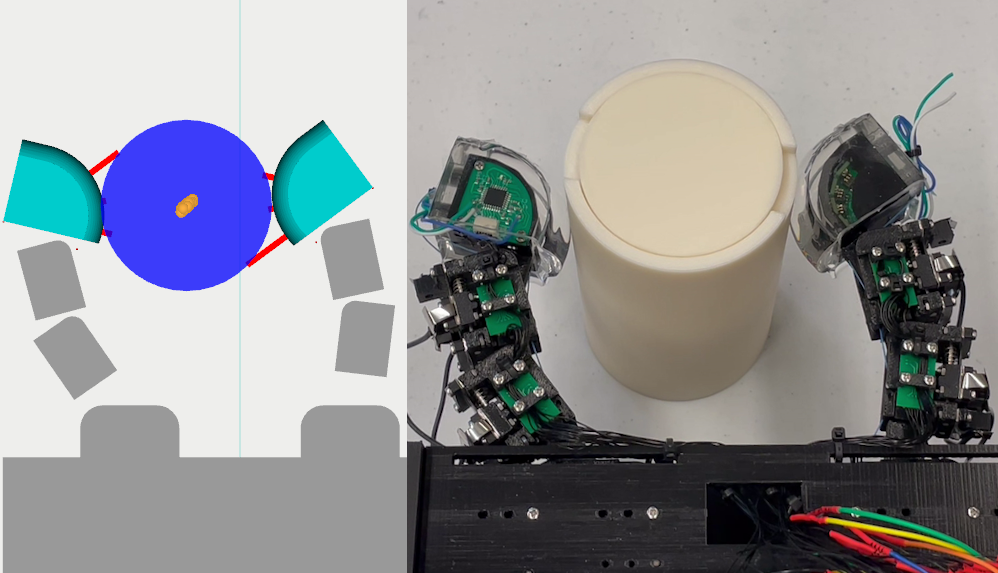}
\caption{Our two fingered hand uses optical proximity sensors to estimate the pose of a cylindrical object during a planar in-hand manipulation task in which the robot must move the object to a specific goal position. Each fingertip is equipped with four optical distance sensing modules that have been covered by a transparent elastomer. Left: A visualization (in RViz) of the robot and object. Red lines emanating from the cyan fingertips illustrate optical distance measurements. A particle filter (particles shown in orange) uses these measurements to estimate the pose of the cylinder shown in blue. Right: The real robot manipulating the object. }
\label{fig:system}
\end{figure}

In-hand manipulation is a prerequisite for many of the tasks that are desirable for robots to perform. Tool use, object articulation, and pick and place tasks will significantly benefit from the ability to reposition, reorient, and exert force on the object of interest within the robot's hand. As the robot attempts to execute a particular in-hand manipulation skill, the appropriate control action at any point in time will depend heavily on the pose of the object. Pose estimation algorithms fuse the robot's control actions and sensor observations across time in order to continuously track the pose of the object. These algorithms must be provided with sensor measurements that disambiguate the pose of the object in order to perform well, but sensing modalities currently used for in-hand manipulation struggle to continuously provide such measurements. Here, we analyze the application of a previously unused sensing modality - fingertip embedded proximity sensors - for \textit{pose estimation during in-hand manipulation} (Fig. \ref{fig:system}). 


The effectiveness of a sensor used for object localization is determined by its ability to measure discriminative features of the object during manipulation. For RGB(D) image sensors, these features are image pixels that correspond to keypoints of the object or a fiducial attached to the object. When these features are occluded from view, pose estimation performance suffers. This is particularly an issue for in-hand manipulation as the fingers will often block environment- or robot-mounted image sensors' view of the object. 

\begin{figure*}
     \centering
     \begin{subfigure}[b]{0.165\textwidth}
         \centering
         \includegraphics[width=\linewidth]{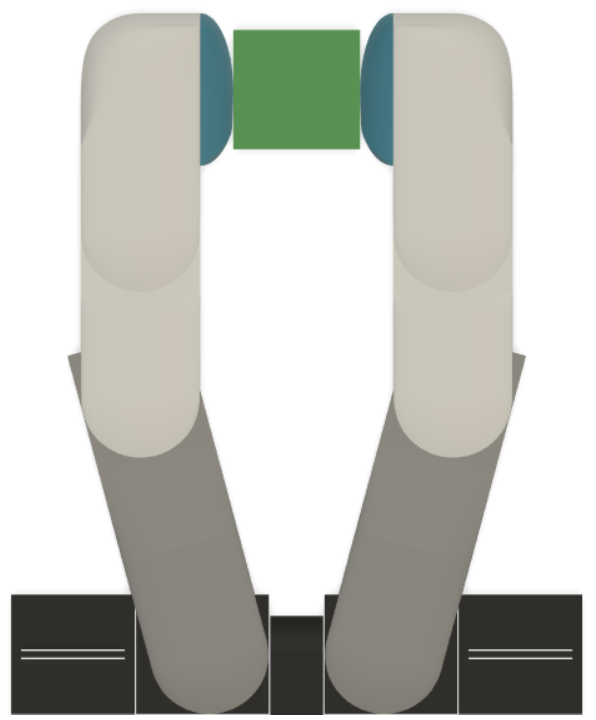}
         \caption{\label{fig:pose_estimation_start}}
     \end{subfigure}
     \hfill
     \begin{subfigure}[b]{0.365\textwidth}
         \centering
         \includegraphics[width=\textwidth]{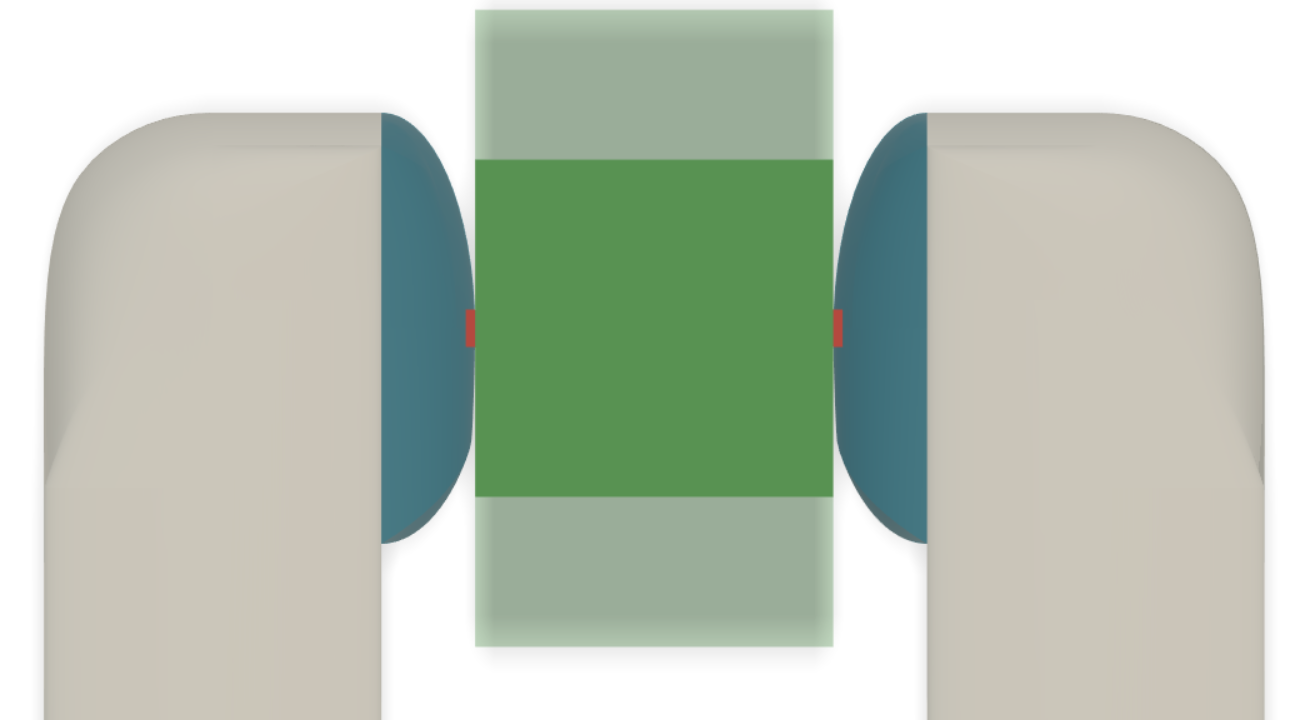}
         \caption{\label{fig:pose_estimation_tactile}}
     \end{subfigure}
     \hfill
     \begin{subfigure}[b]{0.365\textwidth}
         \centering
         \includegraphics[width=\textwidth]{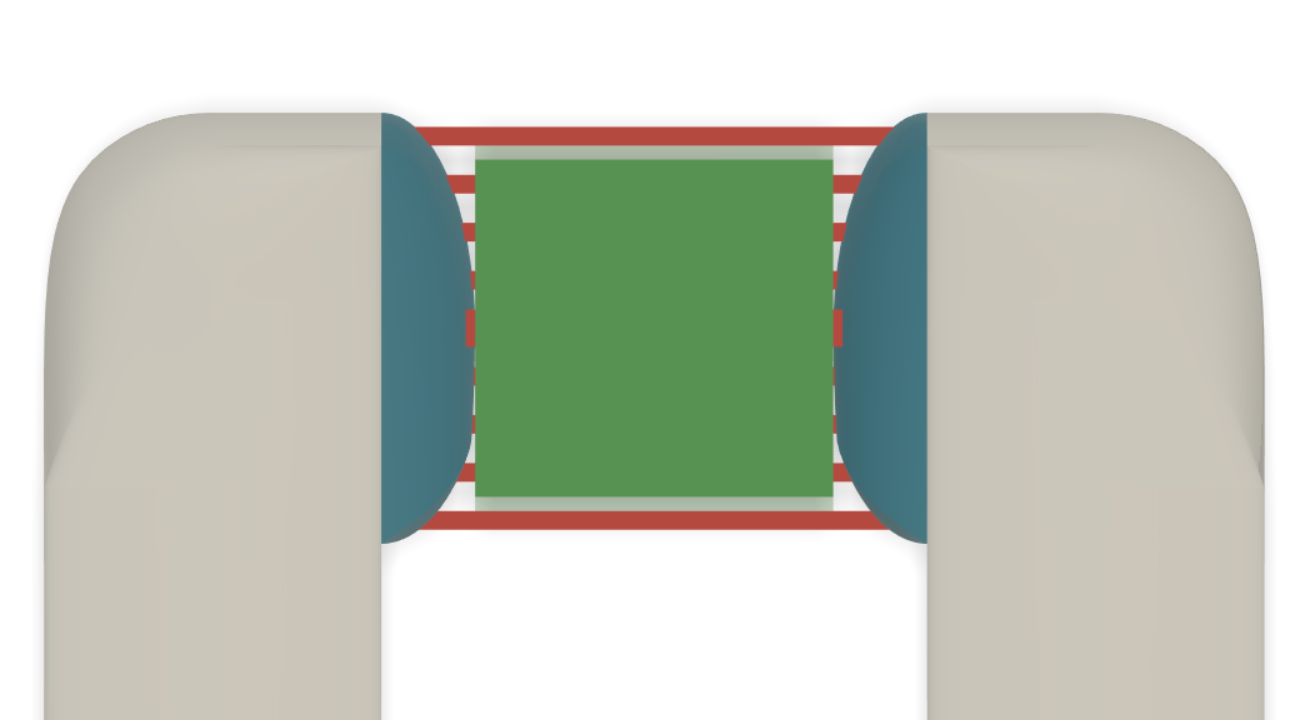}
         \caption{\label{fig:pose_estimation_optical}}
     \end{subfigure}
        \caption{Optical proximity sensors provide measurements beyond the local areas of contact that reduce uncertainty in pose estimation. \subref{fig:pose_estimation_start}) Top down view of a two fingered hand with blue fingertips manipulating a green object. \subref{fig:pose_estimation_tactile}) Zoomed in view in which the red regions indicate object-fingertip contacts that would be detected by tactile sensors. While detection of these contacts results in the object being well localized in the horizontal direction, the translucent green area illustrates the large pose uncertainty in the vertical direction (assuming the geometry of the object is known apriori). \subref{fig:pose_estimation_optical}) Optical distance measurements illustrated by red rays emanating from the fingertips detect a greater portion of the object and thereby significantly reduce pose uncertainty in the vertical direction.}
        \label{fig:pose_estimation}
\end{figure*}

Tactile sensing is the other most commonly used sensing modality applied to in-hand manipulation. Its features consist of contact measurements that are the result of an object's surface exerting force on the sensorized portion of the finger. Tactile sensors can be made robust to the occlusion that image sensors experience by being embedded into the contact surface of the fingers. On the other hand, the ability of tactile sensors to detect discriminative features is limited by the fact that they require contact with the object in order to make measurements. Even if the sensor has a high spatial resolution \cite{yuan2017gelsight,donlon2018gelslim, lambeta2020digit}, objects that do not conform to the sensor's surface (e.g. objects with high amounts of curvature) may be difficult to localize. A sensing modality that improves upon the two described here would be robust to finger occlusions and also be able to measure object geometry beyond the local areas of contact. 

Proximity sensors successfully capture discriminative geometric features for pose estimation where previous sensing modalities fail. The robot's fingers can heavily occlude the object from view of image sensors mounted to the environment or the robot itself, but proximity sensors are robust to such occlusion by virtue of being embedded in the fingers themselves. Tactile sensors embedded in the fingers are also robust to occlusion, but require contact in order to detect the object. Therefore, it can be difficult for tactile sensors to localize objects that do not conform to the sensorized portion of the finger surface. Consider a scenario in which a two fingered hand with slightly rounded fingertips attempts to manipulate an object with flat sides, such as in Fig. \ref{fig:pose_estimation}. In such a scenario, contact detection indicates that the object is within the robot's grasp, but does not elucidate whether the contacts are towards the top corners of the object, bottom corners of the object, or somewhere in between. Proximity sensors can resolve this ambiguity by sensing beyond the local areas of contact, resulting in improved pose estimation.

In this paper, we evaluate the performance of proximity sensor based pose estimation in the context of in-hand manipulation. Specifically, we measure the pose estimation accuracy throughout in-hand manipulation trajectories, and quantify how well a control policy informed by these pose estimates can move the object to a specified goal pose. The insight that proximity sensors are particularly well-suited to pose estimation for in-hand manipulation is rooted in their ability to avoid occlusion by being embedded in robot fingertips while also being able to detect portions of the object that do not necessarily make contact with the fingertip surface. These properties increase the likelihood that the sensing modality will consistently observe discriminative object features throughout the manipulation, facilitating improved pose estimation.\\
\\
\\
We make the following contributions:
\begin{itemize}
    \item Describe a novel two-fingered robot hand for planar object manipulation with optical proximity sensors embedded in its fingertips. 
    \item Present the first demonstration (to the best of the authors' knowledge) of fingertip embedded proximity sensors being used for pose estimation during an in-hand manipulation task.
    \item Compare the accuracy of proximity sensing based pose estimation versus that of tactile based pose estimation.
\end{itemize}
Supplementary video of this work is available here: \url{https://www.youtube.com/watch?v=9iaLuM-N2VE}

\section{Related Work}


This section provides context for our work on proximity sensor based pose estimation for in-hand manipulation. We examine previous works on in-hand manipulation, focusing on the sensing modalities that were used for object pose estimation. We then discuss previous proximity sensors that have been designed for robotic fingertips and their applications.

\subsection{Pose Estimation for In-Hand Manipulation}


Manipulation of an object inherently constrains its pose to some degree. Even without exteroceptive sensing, a robot can reason about how the configuration  of its hand constrains the object in order to estimate its pose with some amount of uncertainty. Sequences  of uncertainty-aware motion primitives allow robots to perform in-hand manipulation without any exteroceptive sensory feedback \cite{bhatt2022surprisingly}. Such information can be combined with constraints imposed by the motions of the arm, gravity, and external contacts to estimate object pose \cite{dafle2014extrinsic, chavan2015prehensile}. While this approach reduces hardware requirements, relying on the executed actions themselves to maintain small pose uncertainty can make it difficult to achieve arbitrary manipulations. Leveraging the constraints on the object's pose imposed by the robot's hand is a useful principle, but the use of exteroceptive sensing helps decouple accurate pose estimation from the particular actions executed by the robot.

The use of image sensors for pose estimation have facilitated the execution of dexterous in-hand manipulations \cite{andrychowicz2020learning,bircher2021complex,hang2021manipulation,morgan2022complex, higo2018rubik}. This approach generally requires object mounted visual fiducials and/or a spatially distributed array of cameras. Other environments may not be able to fulfill such requirements, resulting in occlusion and degraded performance. 

On the other hand, fingertip embedded tactile sensors are robust to occlusion. Maekawa et al. \cite{maekawa1995tactile} demonstrated one of the earliest works in manipulating objects with rolling contacts along a desired trajectory using tactile feedback at the fingertips. More recent works have used tactile sensors to learn model-free control policies for in-hand manipulation \cite{kumar2016learning, van2015learning, falco2018policy}. Others have focused on using tactile measurements for explicit pose estimation \cite{liang2020hand, koval2015pose, zhang2012application, bauza2020tactile}. However, tactile based pose estimation may struggle to localize objects that do not conform to the surface of the fingertip. 

%



\subsection{Proximity Sensing}

Proximity sensors provide a sense of “pretouch,” with a range intermediate to that of image and tactile sensors. We classify prior work on the proximity sensing modality for robot hands based on the underlying sensing mechanism: either acoustic, capacitive, or optical. While we describe many works here, please see Navarro et al. \cite{navarro2021proximity} for a more comprehensive survey of proximity sensing in robotics.

Jiang and Smith \cite{jiang2012seashell} develop acoustic pretouch sensors embedded in the fingertips of a PR2 robot. They  measure proximity to objects by detecting changes in the ambient noise spectrum inside of the sensor's acoustic cavity. One disadvantage is that the sensor fails to detect extermely soft and light materials. Fang et al. \cite{fang2019toward} propose a bi-modal acoustic-optical sensor using the optoacoustic effect.

Capacitive sensors extract information from the environment by measuring changes in capacitance between two or more sensor electrodes. Such sensors can operate in both tactile and proximity modes \cite{goger2013tactile}. Wistort et al. \cite{wistort2008electric} use electric field pretouch as the feedback signal for closed loop control in robotic manipulation tasks. Mayton et al. \cite{mayton2010electric} extend this principle for co-manipulation of objects between humans and robots. Faller et al. \cite{faller2019design} retrofit an industrial gripper system with capacitive sensors for automated grasping of logs in a forestry robot. Muhlbacher-Karre \cite{muhlbacher2015responsive} integrate capacitive sensing for active object categorization in robot manipulation tasks. However, the performance of capacitive sensors degrades when sensing objects with a low dielectric constant such as fabrics, foam, and plastics.

Optical pretouch sensing methods are attractive because of their precision and ability to detect a wide range of materials. Guo et al. \cite{guo2015transmissive} proposed to perceive objects using a transmissive optical proximity sensor composed of an emitter and a receiver in a parallel gripper. Hsiao et al. \cite{hsiao2009reactive} use optical sensors for pose estimation using a probabilistic model. Maldonado et al.\cite{maldonado2012improving} augment long-range vision with optical sensors to obtain measurements of areas that are occluded from the image sensor. 

Yang et al. \cite{yang2017pre} demonstrate continuous manipulation tasks using a time-of-flight (ToF) proximity sensor mounted on a parallel gripper. 
Other works have developed either camera or optical sensors encased in transparent, compliant materials that facilitate both proximity and tactile sensing during manipulation \cite{patel2016integrated, lancaster2019improved, yamaguchi2017implementing}. Sasaki et al. \cite{sasaki2018robotic} exploit ToF sensors to more robustly grasp objects.  Optical sensors may fail to detect objects that are transparent or highly specular. Methods that have been proposed to compensate for variations in reflectance include color calibration information obtained by a vision sensor \cite{konstantinova2016fingertip} and using a light-emitting diode with emission phase that is robust to varying reflectance \cite{koyama2019high}.


\section{Proximity Sensing for In-Hand Manipulation}

To investigate the value of proximity sensing for in-hand manipulation, we deploy a complete system for object manipulation under pose uncertainty on our custom two-fingered robot hand. We first present the overall robot hardware architecture, and then focus on our design of compliant fingertips with embedded optical time-of-flight proximity sensors. We then describe our implementation of a particle filter that uses measurements from these sensors to estimate the object's pose, and conclude by detailing the model predictive control policy that our robot used to perform in-hand manipulation.

\subsection{Two-Fingered Robot for In-Hand Manipulation}

We developed a two-fingered robot hand capable of in-hand manipulation in order to explore the use of proximity sensors for pose estimation (Fig. \ref{fig:system}). For each finger, a tendon (consisting of 65 lb. max tension fishing line) is anchored to the fingertip and then routed through the links of the finger in order to be attached to a XM430-W350-R Dynamixel servo motor. Flexion of a finger occurs when its corresponding motor pulls on the tendon, and springs embedded in each joint extend the finger as the motor releases the tendon. Each finger contains three revolute joints. 

Although the fingers are underactuated, a unique feature of the robot hand is that the rotation of each individual joint can be blocked by a corresponding electrostatic brake, allowing the motion of the joint to be decoupled from the motor as desired. Electrostatic brakes leverage the electrostatic attraction that occurs between two conductors at differing voltage potentials. This attraction induces a perpendicular frictional force that resists motion between the conductors. Our electrostatic brake equipped joint uses a rack and pinion to transform rotational motion of the joint into linear sliding between the conductors, which optimizes conductor conformance in order to achieve significant braking capability. More details of this joint design are available in \cite{lancaster2022braking}; for this work, one only needs to note that the robot can engage any combination of joints' brakes in order to block the motion of those joints. These brakes enable the robot to control the motion of individual joints, allowing the hand to reach any arbitrary joint configuration within the joint limits.

Each joint contains an AEAT-8800 magnetic encoder. The encoders measure the angular position of each joint at a rate of 30 Hz. The limits for each joint are from 0 to 90 degrees.   

\subsection{Compliant Fingertips with Embedded Proximity Sensors}

Optical time-of-flight sensors possess a number of qualities that facilitate the practical implementation of proximity sensor based pose estimation. This type of sensor makes easily interpretable (i.e. distance) measurements that are insensitive to most object compositions and surface properties. The ability to compactly manufacture such sensors allows them to be embedded in compliant regions of the finger, and this compliance aids the robot hand in maintaining a grasp on the object. This subsection describes the design and construction of our robot hand's sensorized fingertips.

\begin{figure}[t]
\centering
\includegraphics[width=0.45\textwidth]{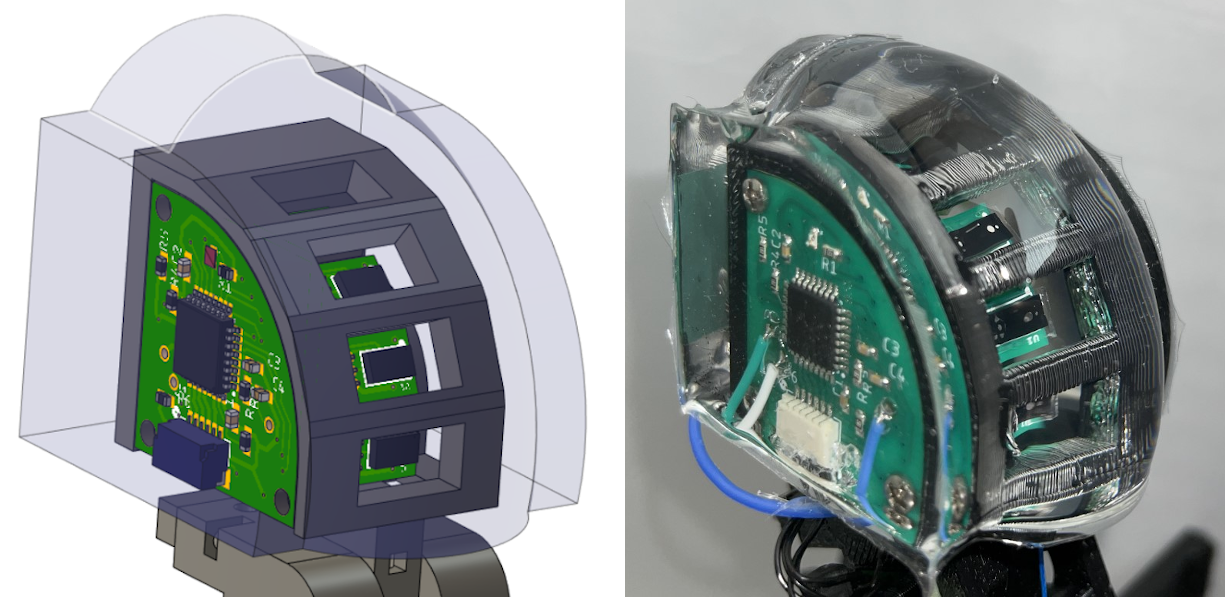}
\caption{A compliant fingertip with four embedded optical time-of-flight proximity sensors. Left: A visualization of the fingertip design. PDMS (colored in translucent blue) forms a compliant contact surface. Right: The constructed fingertip.}
\label{fig:fingertip}
\end{figure}

Four STMicrolelectronics VL6180x optical time-of-flight modules are embedded into each of the robot's fingertips as shown in Fig. \ref{fig:fingertip}. The sensor modules of each fingertip are positioned along a circular arc of radius 15mm at intervals of 25 degrees, and oriented such that their optical transmitters point perpendicular to the arc's tangent. Each sensing module produces distance measurements at a rate of 30 Hz, which are sent to a fingertip embedded microcontroller that provides the measurements to the robot's main computer via the same communication protocol used to query the encoders.

After the fingertip's embedded electronics are mounted to a 3D printed substrate, this rigid skeleton is covered by transparent, compliant polydimethylsiloxane (PDMS). We design a mold for the fingertip that fulfills two critera:
\begin{itemize}
    \item For each sensor module, the contact surface above the module's optical transmitter and receiver should be shaped as a circular arc centered around the transmitter in order to focus infrared light reflecting off of the inner contact surface back towards the transmitter (as opposed to towards the receiver). This results in better contrast-to-noise ratio performance of the sensor.
    \item The contact surface should be at a distance above the 10mm minimum sensing range of the module (here, we use a distance of 15mm).
\end{itemize}

Once a satisfactory mold has been 3D printed, the surface is covered with two coats of Smooth-On XTC3D Epoxy. This epoxy fills in the gaps between the individual 3D printed layers and is necessary to produce a contact surface that is optically transparent. The mold is then filled with a 10:1 silicone-catalyst PDMS mix, thoroughly degassed using a vaccum chamber and pump, and cured for 24 hours. For more details on this process, please see our previous work on the construction of optical proximity-contact-force sensors \cite{lancaster2019improved}. 

\subsection{Proximity Sensor Based Pose Estimation}
\label{sec:pf}
By virtue of using similar measurement mechanisms, many of the standard techniques applied to LIDAR based localization in the context of mobile robots can be adapted for proximity sensor based object pose estimation \cite{thrun2001robust}. In particular, we implement a particle filter for estimating the pose of the manipulated object. Our encoder measurements themselves provide a sufficiently accurate estimate of the robot's internal state, therefore our particle filter's estimated state $x_t$ only consists of the cylindrical object's xy position.

Given $N$ particles, our particle filter's sensor model generates particle weights $p(z_t|x^i_t)$ that represent the likelihood of sensor measurements $z_t$ occurring given each particle hypothesis $x^i_t$, where $i \in \{1,\dots, N\}$. Like other beam-based sensor models, we assume that each individual sensor beam measurement is conditionally independent of all of the others given the underlying state:

\begin{equation}
     p(z_t|x^i_t) = \prod^M_{j=1} p(z^j_t|x^i_t)
\end{equation}
with $M = 8$ due to the robot having two fingers, each with four sensing modules (see Fig.~\ref{fig:fingertip}). The individual observation probabilities are a weighted mixture of a Gaussian distribution representing the expected distance measurement with noise and a uniform distribution representing the possibility of a sensor glitch, communication error, unexpected detection, or other random effects:

\begin{equation}
     p(z^j_t|x^i_t) = \eta \cdot \left(w_1 \cdot \mathcal{N}(z^{j*}_t ,\,\sigma^{2}) + w_2 \cdot \mathcal{U}(0,z_{max})\right)
\end{equation}
where $z^{j*}_t$ is the expected measurement of sensor $j$ given the current encoder values, robot forward kinematics, and particle hypothesis $x^i_t$. Here, $w_1$ and $w_2$ are mixing weights chosen according to the importance of the corresponding term, and $\eta$ serves as a normalizer to ensure that the sum over the distribution is one.

Each time the robot executes an action, our particle filter's motion model propagates its particles according to the expected change in position plus zero-mean gaussian noise. The expected change in position is computed by executing the action in a physics simulator for robotics. This simulator is a key component of our model predictive control policy, and is discussed in further detail in the following subsection.  

\begin{figure*}[t]
\centering
\includegraphics[width=\textwidth]{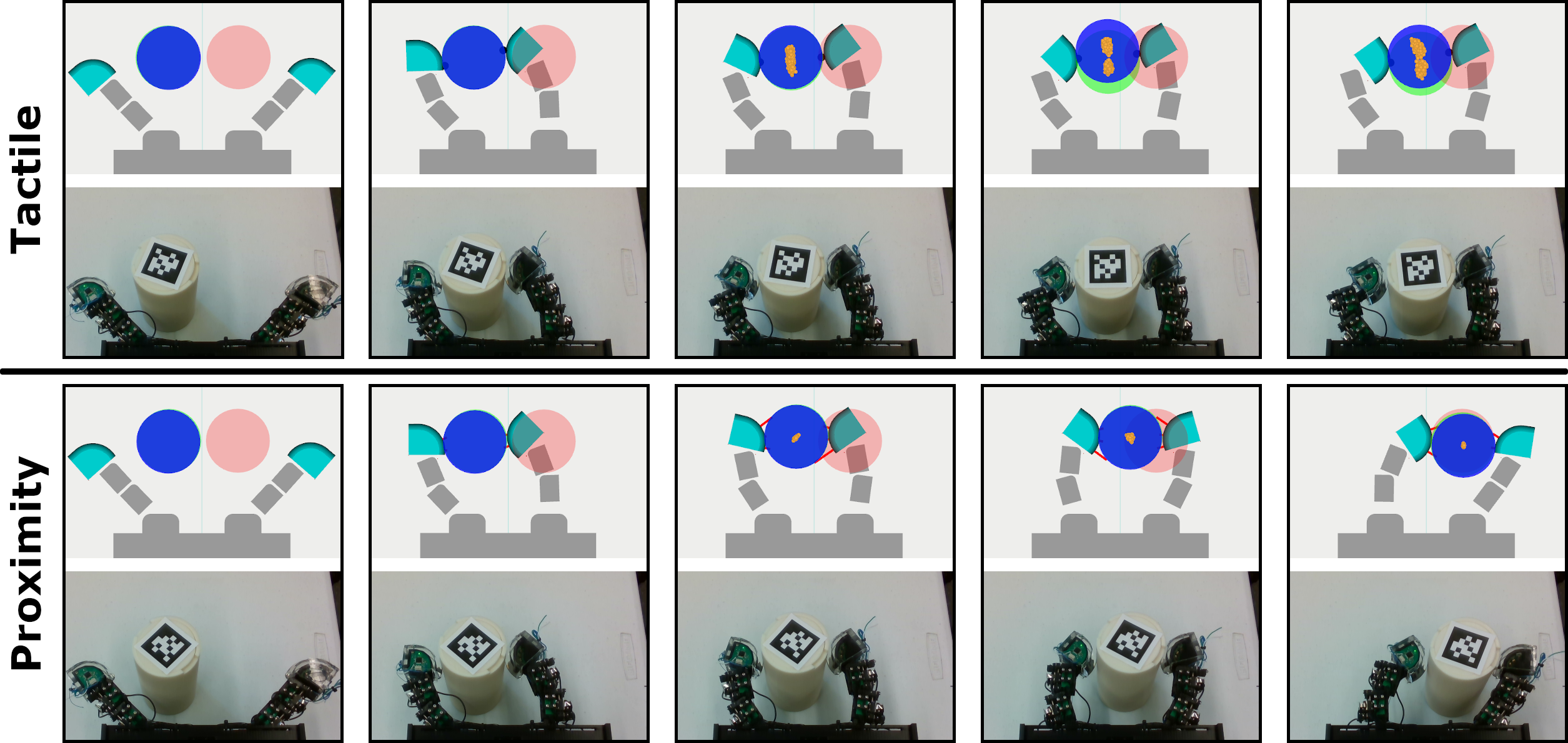}
\caption{ Representative in-hand manipulation trials using tactile sensor and proximity sensor based pose estimation. The blue cylinder represents the object pose estimated by the particle filter (particles shown in orange). The green cylinder (heavily overlapped by the blue cylinder, and most visible in images to the upper right) represents the ground-truth pose. The red cylinder represents the desired goal pose of the object. Note that the visual fiducials are only used to estimate the ground-truth pose; they do not affect the sensor based pose estimation nor the executed control policy. The left-most snapshots show the initial pose for both trials, and the snapshots second from the left show when the cyan fingertips initially make contact with the object. Subsequent snapshots show the progression of the manipulations. Top: A tactile sensor manipulation trial. In the upper images, black spheres along the fingertip surface illustrate detected contacts. Here, pose estimation error pushes the system into an area of the state space in which the controller can no longer make progress towards the goal. Bottom: A proximity sensor manipulation trial. Red lines emanating from the fingertips illustrate proximity sensor measurements.
}
\label{fig:optical_traj}
\end{figure*}

\subsection{Model Predictive Control for In-Hand Manipulation}
\label{sec:mpc}
We use model predictive control (MPC) to generate actions for in-hand manipulation. Starting with the current system state, MPC simulates a large number of action sequences in order to find trajectories that will move the system towards the goal state over the trajectories' time horizon. By repeating this process at each timestep, MPC provides adaptive control that is robust to unmodeled dynamics.


Given the current state $s_t$ of the robot-object system and the action $a_t$ that the robot executes, our dynamics model predicts the resulting next state $s_{t+1}$. The state $s_t$ consists of the six joint positions and velocities, as well as the $xy$ position and velocity of the manipulation object. The action $a_t$ is composed of the commanded positions of the robot's two motors, and the states of each of the six brakes (on or off). While in general any combination of brakes can be turned on, for this work we limit executed braking configurations to be those in which exactly one of the brakes in each finger is off. By reducing the number of possible braking configurations from 64 to 9, it is more tractable for the model predictive controller to explore the action space. We use NVIDIA's Isaac Gym \cite{makoviychuk2021isaac}, a  GPU based physics simulator, as a highly parallelizable dynamics model.

Our in-hand manipulation controller is an adapted version of the model predictive path integral (MPPI) framework \cite{williams2017information, zhong2019}. The standard MPPI controller generates a large number of action sequences, simulates those sequences to obtain corresponding trajectories and costs, and then outputs a cost-weighted average of those action sequences for execution on the real robot. However, this averaging does not make sense for a hybrid action space for which the discrete variables lack a Euclidean distance measure. Instead, we require each action sequence to maintain a consistent brake configuration throughout the corresponding simulated trajectory, and then compute cost-weighted averaged action sequences for each possible brake configuration. Initially, of the 9 outputted action sequences, we execute the first action of whichever sequence has the lowest cost. For subsequent time steps, we only choose an action sequence corresponding to a different brake configuration if it has a significantly lower cost (we use a constant threshold percentage $\phi$) than the cost of the action sequence corresponding to the previous brake configuration. Otherwise, we execute the first action of the action sequence corresponding to the previous brake configuration. 

At each time step, our MPPI controller simulates many trajectories over a time horizon $T$. It computes a cost for each trajectory that penalizes the fingers not making contact with the object and object distance from the goal:

\begin{equation}
    J(s_t, \dots, s_{t+\tau}) = c_1 \cdot \displaystyle \sum_{\tau' = t}^{t+\tau}  \mathcal{I}(s_{\tau'}) + c_2 \cdot \lvert x_{goal}-x_{t+\tau} \rvert 
\end{equation}
where $\mathcal{I}(s_{\tau'}) $ is an indicator function that returns the number of fingertips \textit{not} in contact with the object, and $x_{goal}$ is the desired position of the object.

\section{Evaluation}
\label{sec:results}

We measure the performance of pose estimation for in-hand manipulation with different sensing modalities. This section discusses our experiment's specifications and results.

\subsection{Experiment Setup}

We consider a planar in-hand manipulation task in which the robot must translate an object from a known initial location to a goal location. These locations are chosen to be on the boundary of the robot's manipulation workspace. The object's initial pose is 4.5cm to the left of the geometric plane that symmetrically bisects the hand (see Fig.~\ref{fig:optical_traj}), and 4.5cm above the base of the fingers. The goal pose is the reflection of the initial pose across the bisecting plane. The object is a 3D printed cylinder of radius 4cm and a height of 14cm. 

When attempting the in-hand manipulation task, the robot begins at a position in which all joints have a value of zero (Fig.~\ref{fig:optical_traj} on the left). It then uses a joint position controller to move to a preset pose that corresponds to the robot making initial contact with the object at the known initial pose. Once contact has been made, the MPPI controller attempts to move the object to the goal pose. A successful manipulation trial ends when the horizontal distance from the object's ground-truth location (provided by visual fiducial localization) to the goal location is less than 1mm. A trial ends and is considered to have failed if the previous condition is not fulfilled within 60 seconds after first making contact with the object. 

We measure the robot's in-hand manipulation performance when using pose estimation based on three different sensing modalities. These three modalities are tactile sensing, proximity sensing, and visual fiducial localization. For each of the sensing modalities, we measure the distance between the goal pose and the object's pose at the end of the manipulation. Fiducial localization also serves as a ground-truth pose estimate during the tactile and proximity sensor trials in order to estimate pose estimation accuracy throughout the manipulation. Ten manipulation trials are undertaken for each of the three sensing modalities.

We implement tactile sensing by artificially limiting the range of the embedded proximity sensors to be just beyond the contact surface of the fingertip. Although the sensors can detect objects up to 255mm away, a tactile sensor would only be able to detect objects at a distance less than or equal to the distance from the sensor to the finger surface $d_{tact\_max}$. Therefore, all measurements greater than $d_{tact\_max}$ are truncated to $d_{tact\_max}$ when in tactile sensing mode. We perform this truncation for both the real observations $z^{j}_t$ and expected observations $z^{j*}_t$ of the particle filter.

\subsection{Implementation Details}

The algorithms described in Sections \ref{sec:pf} \& \ref{sec:mpc} are simultaneously executed on a single desktop PC with an Intel i7 Quad-Core CPU, 64 GB of RAM, and a NVIDIA Titan XP GPU. Initial values for all of the following parameters were chosen based on our intuition and then hand-tuned until reasonable performance was achieved. Our particle filter used 1000 particles and produced pose estimates at a rate of 18 Hz. Its sensor model used parameter values $\sigma = 5mm$, $w1 = 0.95$, $w2 = 0.05$. Although the distance from any sensor to the fingertip surface is 15mm, we set $d_{tact\_max}$ to a slightly larger 18mm to ensure that sensor noise does not result in false negatives. Our MPPI controller simulates 297 trajectories over a time-horizon $T=10$ at each time step. Its cost function uses parameter values $c_1 = 0.1$, $c_2 = 200$, and the MPPI hyperparameter $\lambda$ is set to 0.1. Switching between braking configurations is thresholded on a value of $\phi = 25\%$. Controls are generated at a rate of 5 Hz.

\subsection{In-Hand Manipulation Performance}

Relative to tactile sensor based pose estimation, we found that proximity sensor based pose estimation is more accurate throughout the trajectory and results in more precise positioning of the object. We observed that the spread of the particle distribution in the vertical direction is larger for tactile sensor based pose estimation (Fig. \ref{fig:optical_traj}). This is exemplified by the second-to-last snapshot of the tactile sensing sequence in which the distribution has bifurcated around two distinct hypotheses. This may be due to the sensor data not being sufficiently discriminative to determine whether the robot is grasping above or below the diameter of the cylinder.

\begin{figure}
     \centering
     \begin{subfigure}[b]{0.24\columnwidth}
         \centering
         \includegraphics[width=\columnwidth]{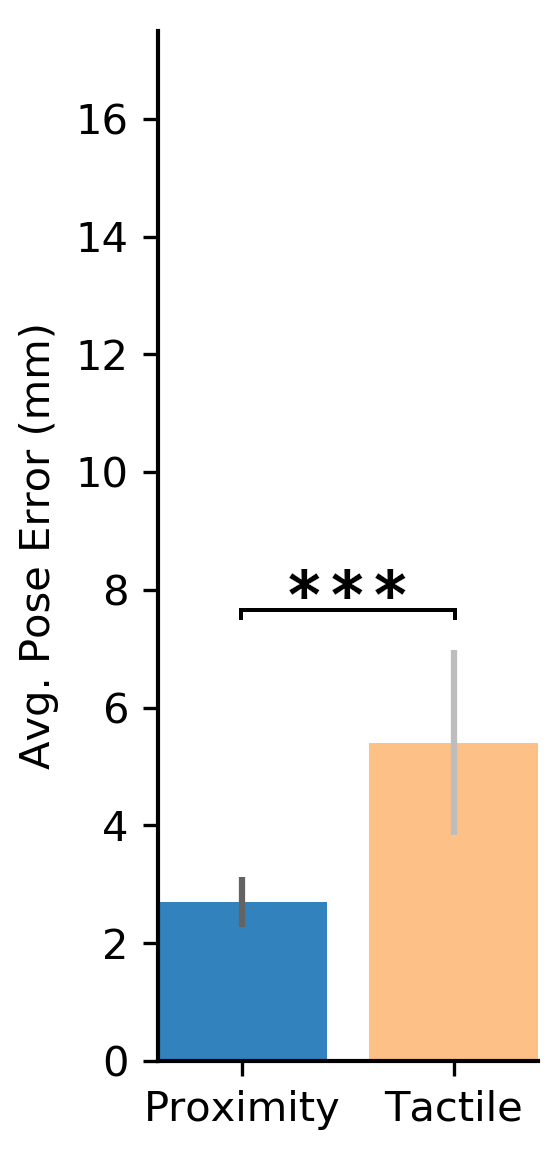}
         \caption{\label{fig:avg_pose_error}}
     \end{subfigure}
     \hfill
     \begin{subfigure}[b]{0.33\columnwidth}
         \centering
         \includegraphics[width=\columnwidth]{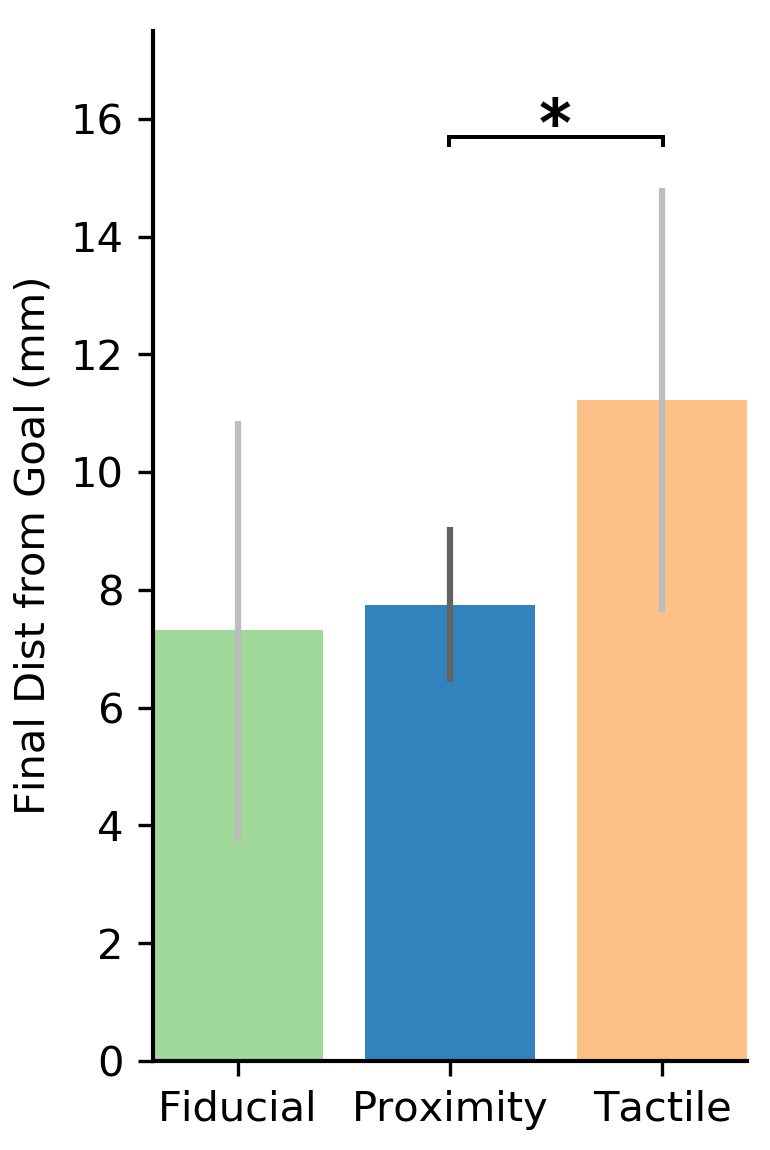}
         \caption{\label{fig:goal_dist}}
     \end{subfigure}
     \hfill
     \begin{subfigure}[b]{0.33\columnwidth}
         \centering
         \includegraphics[width=\columnwidth]{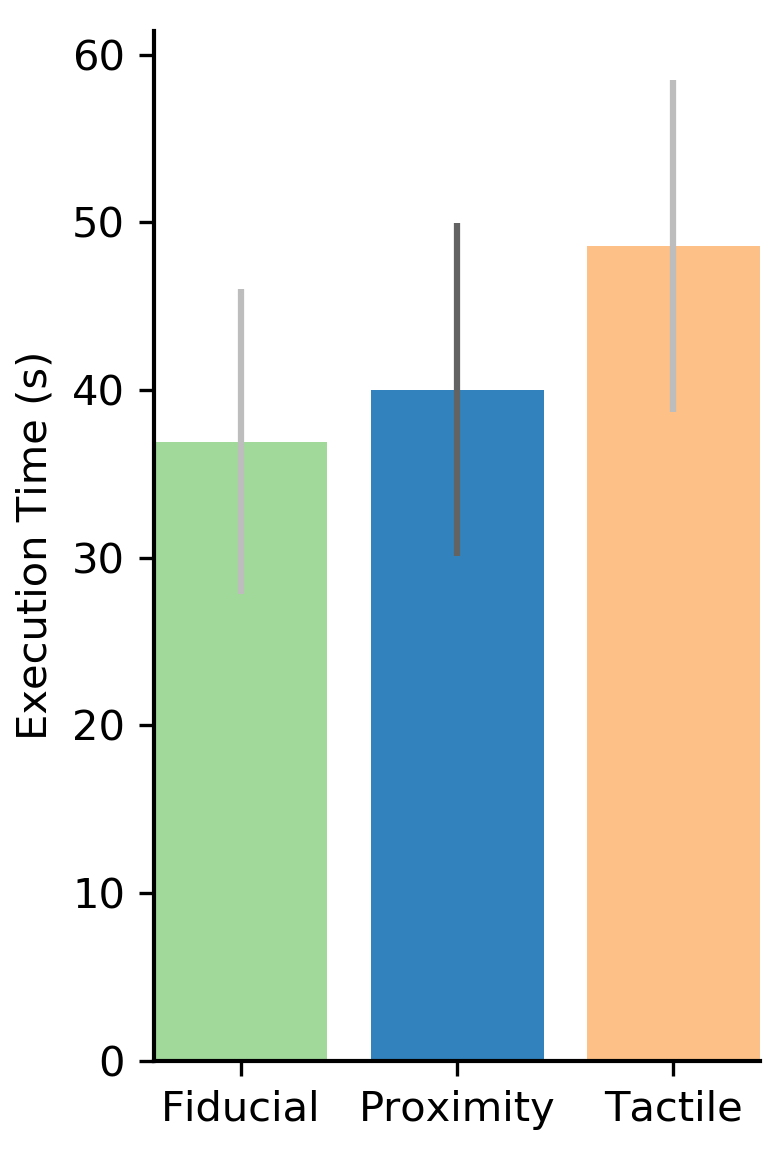}
         \caption{\label{fig:exec_time}}
     \end{subfigure}
     \caption{Comparison between pose estimation sensing modalities. Error bars correspond to one standard deviation. Increasing number of stars indicates higher significance levels ($p < 0.05$, $p < 0.01$, $p < 0.001$) according to a paired U test. \subref{fig:avg_pose_error}) The average pose error. \subref{fig:goal_dist}) The object's final distance from the goal. \subref{fig:exec_time}) Time between initial contact and task completion.}
     \label{fig:traj_results}     

\end{figure}

Across our experiments, we found that proximity sensing enables better performance than tactile sensing. Proximity-sensor based pose estimation achieves significantly lower average pose error than tactile-sensor based pose estimation ($p < 0.001$, Mann-Whitney U test). The average pose error throughout the manipulation for proximity sensing was 2.7mm, less than half of the 5.5mm error observed for tactile sensing (Fig. \ref{fig:traj_results}). The quality of pose estimation is also reflected in the overall task performance. When using visual fiducial or proximity sensor based pose estimation, the robot was able to successfully complete all ten trials. However, only five out of ten trials were successful when using tactile-based pose estimation. 
For the five trials that succeeded, tactile sensing based pose estimation trials resulted in 11.2mm average final position error. In contrast, proximity-sensor based pose estimation trials resulted in an average final positioning error of 7.8mm across all trials, 30\% lower than that of tactile sensing, enabling also significantly lower final positioning error compared to tactile sensing ($p < 0.05$, U test). Furthermore, proximity-sensor based pose estimation resulted in an average task completion time of 40.0 seconds, 18\% faster than the 48.6 second average completion time of (successful) tactile sensing trials. 
The complete statistics for all experiments are listed in Table~\ref{tab:results}.

\begin{figure}[t]
\centering
\includegraphics[width=0.45\textwidth]{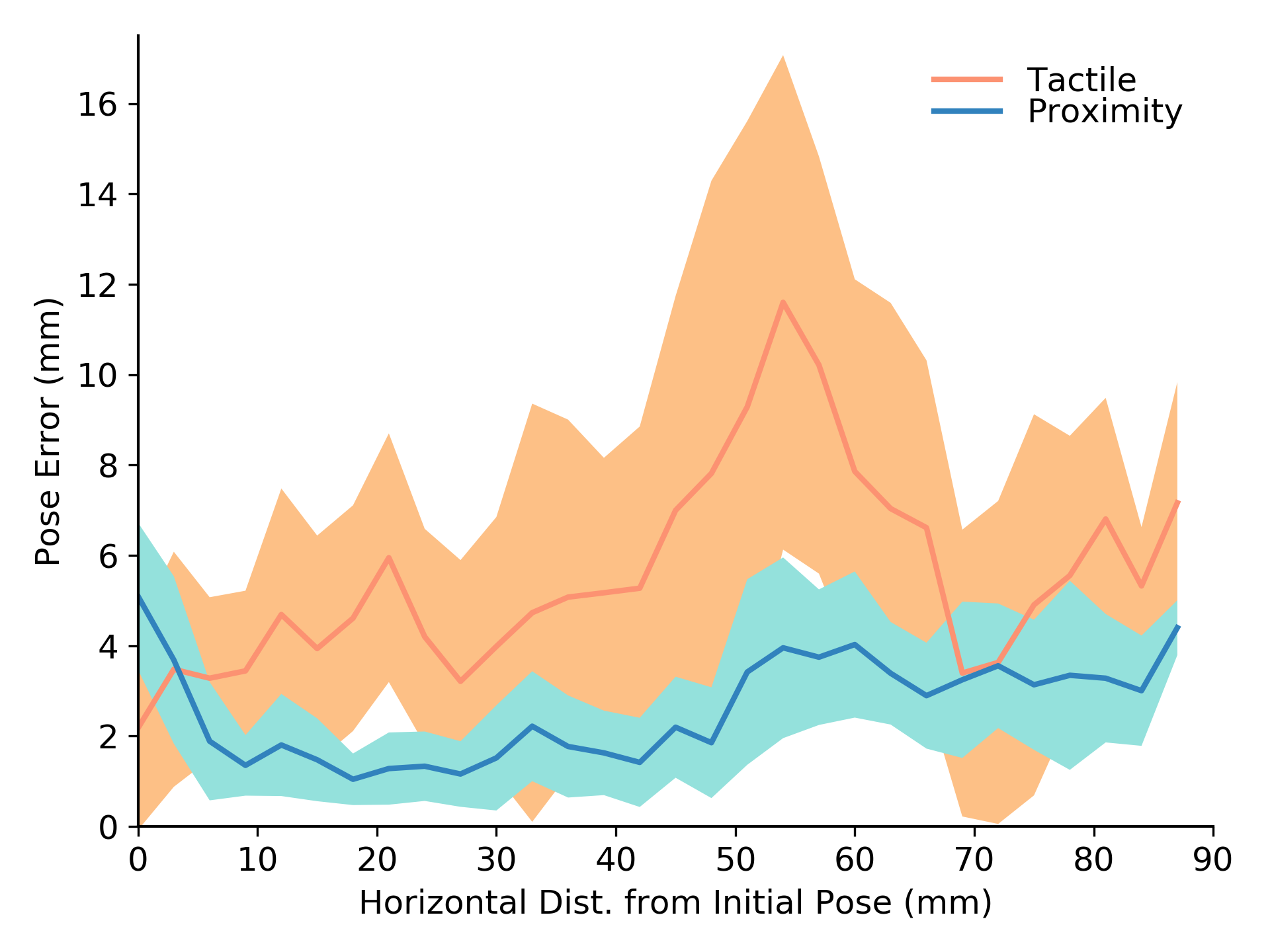}
\caption{The error in pose estimation for both tactile and proximity sensing. Each series is averaged over ten trials, and the shaded errors represent one standard deviation.}
\label{fig:traj_error}
\end{figure}

It is also informative to examine how the pose error evolves throughout the in-hand manipulation task. With the exception of the very beginning of the task, proximity-sensor based pose estimation has lower error than tactile based pose estimation as shown in Fig.~\ref{fig:traj_error}. We hypothesize that the tactile modality does better at the beginning due to a mismatch between the particle filter's sensor model and the actual sensor. The optical beams of the sensor model are represented by beams with an infinitesimally small width, but the real sensor's detection zone takes the shape of a cone. For the particular configuration in which the robot initially makes contact with the object, the cylinder is just barely within the conic field-of-view of one of the sensors. This causes the sensor to return a measurement significantly less than its max range, but the corresponding sensor model expected measurement is at max range (because the beam has no width). Tactile-based pose estimation is better able to filter this out because its (truncated) max range is much closer to the measurement returned by the actual sensor. However, because the true sensor's detection cone is quite thin (with a radius on the scale of millimeters depending on the distance from the sensor), this configuration in which the object barely grazes one of the sensing cones does not occur often throughout the overall manipulation. Once the particle filter has coalesced around the true state by observing measurements outside of this configuration (i.e. the manipulation proceeds), it naturally blunts the effect of this type of rare configuration by conditioning upon the previous state distribution.

\begin{table}[t]
\centering
\caption{The results of ten in-hand manipulation trials.}
\label{tab:results}
\resizebox{\columnwidth}{!}{%
\begin{tabular}{lcccc}\toprule

         & Avg. Pose Err. (mm) & Success & Goal Dist (mm) & Exec. Time (s)\\
\midrule
Fiducial &  -                  & 10/10   & 7.3  $\pm$ 3.6 & 36.9  $\pm$ 9.1\\
Proximity  & 2.7 $\pm$ 0.4       & 10/10   & 7.8  $\pm$ 1.3 & 40.0  $\pm$ 10.0 \\
Tactile  & 5.5 $\pm$ 1.6       & 5/10    & 11.2 $\pm$ 3.6 & 48.6  $\pm$ 10.0 \\
\bottomrule

\end{tabular}
}
\end{table}

\section{Conclusion}

In this paper, we measured the accuracy of tactile sensor and proximity sensor based pose estimation and its effect on in-hand manipulation performance. The novel application of proximity sensing to in-hand manipulation is motivated by the fact that fingertip-embedded proximity sensors are robust to occlusion (unlike environment- or robot-mounted image sensors) and able to sense beyond the local areas of contact (unlike tactile sensors). We considered an in-hand manipulation task that involved moving an object from one side of its workspace to the other. Our experiments demonstrate that proximity sensor based pose estimation has an average error of only 2.7 millimeters, which is less than half of that of tactile sensor based pose estimation. In terms of task performance, proximity sensing resulted in significantly more robust task completion and final object positioning accuracy relative to tactile sensing.

\subsection{Limitations}

One direction for future work is to perform similar experiments for a wider range of in-hand manipulation tasks. In particular, these experiments could be expanded by using manipulation objects of different size, weight, and geometric shape. The task could be further generalized by using arbitrary initial and goal poses within the robot's workspace.

A limitation of our work is the way in which we implemented tactile sensing. Particularly, higher resolution tactile sensors will likely perform better than our spatially sparse tactile sensor baseline. Although we directly compare tactile sensing against proximity sensing for pose estimation in our experiments, there are other aspects of manipulation that require some form of tactile sensing. Specifically, tactile sensors can detect applied forces, object slippage, vibrations, etc. that can be useful for manipulation but would be difficult or impossible to measure with purely proximity sensors. Ultimately, we believe that future work should focus on developing fingertip sensors that use either new sensing technologies or a mix of existing mechanisms to achieve high resolution proximity \textbf{and} tactile sensing.




\section*{ACKNOWLEDGMENTS}

This work was funded by the National Science Foundation EFRI (\#1832795), National Science Foundation IIS (\#2007011), National Science Foundation DMS (\#1839371), the Office of Naval Research, US Army Research Laboratory CCDC, Amazon, and Honda Research Institute USA.

\balance
\bibliography{bib/IEEEabrv.bib,bib/IEEEexample.bib}{}

\begin{thebibliography}{10}
\providecommand{\url}[1]{#1}
\csname url@samestyle\endcsname
\providecommand{\newblock}{\relax}
\providecommand{\bibinfo}[2]{#2}
\providecommand{\BIBentrySTDinterwordspacing}{\spaceskip=0pt\relax}
\providecommand{\BIBentryALTinterwordstretchfactor}{4}
\providecommand{\BIBentryALTinterwordspacing}{\spaceskip=\fontdimen2\font plus
\BIBentryALTinterwordstretchfactor\fontdimen3\font minus
  \fontdimen4\font\relax}
\providecommand{\BIBforeignlanguage}[2]{{%
\expandafter\ifx\csname l@#1\endcsname\relax
\typeout{** WARNING: IEEEtran.bst: No hyphenation pattern has been}%
\typeout{** loaded for the language `#1'. Using the pattern for}%
\typeout{** the default language instead.}%
\else
\language=\csname l@#1\endcsname
\fi
#2}}
\providecommand{\BIBdecl}{\relax}
\BIBdecl

\bibitem{yuan2017gelsight}
W.~Yuan, S.~Dong, and E.~H. Adelson, ``Gelsight: High-resolution robot tactile
  sensors for estimating geometry and force,'' \emph{Sensors}, vol.~17, no.~12,
  p. 2762, 2017.

\bibitem{donlon2018gelslim}
E.~Donlon, S.~Dong, M.~Liu, J.~Li, E.~Adelson, and A.~Rodriguez, ``Gelslim: A
  high-resolution, compact, robust, and calibrated tactile-sensing finger,'' in
  \emph{2018 IEEE/RSJ International Conference on Intelligent Robots and
  Systems (IROS)}.\hskip 1em plus 0.5em minus 0.4em\relax IEEE, 2018, pp.
  1927--1934.

\bibitem{lambeta2020digit}
M.~Lambeta, P.-W. Chou, S.~Tian, B.~Yang, B.~Maloon, V.~R. Most, D.~Stroud,
  R.~Santos, A.~Byagowi, G.~Kammerer \emph{et~al.}, ``Digit: A novel design for
  a low-cost compact high-resolution tactile sensor with application to in-hand
  manipulation,'' \emph{IEEE Robotics and Automation Letters}, vol.~5, no.~3,
  pp. 3838--3845, 2020.

\bibitem{bhatt2022surprisingly}
A.~Bhatt, A.~Sieler, S.~Puhlmann, and O.~Brock, ``Surprisingly robust in-hand
  manipulation: An empirical study,'' \emph{arXiv preprint arXiv:2201.11503},
  2022.

\bibitem{dafle2014extrinsic}
N.~C. Dafle, A.~Rodriguez, R.~Paolini, B.~Tang, S.~S. Srinivasa, M.~Erdmann,
  M.~T. Mason, I.~Lundberg, H.~Staab, and T.~Fuhlbrigge, ``Extrinsic dexterity:
  In-hand manipulation with external forces,'' in \emph{2014 IEEE International
  Conference on Robotics and Automation (ICRA)}.\hskip 1em plus 0.5em minus
  0.4em\relax IEEE, 2014, pp. 1578--1585.

\bibitem{chavan2015prehensile}
N.~Chavan-Dafle and A.~Rodriguez, ``Prehensile pushing: In-hand manipulation
  with push-primitives,'' in \emph{2015 IEEE/RSJ International Conference on
  Intelligent Robots and Systems (IROS)}.\hskip 1em plus 0.5em minus
  0.4em\relax IEEE, 2015.

\bibitem{andrychowicz2020learning}
O.~M. Andrychowicz, B.~Baker, M.~Chociej, R.~Jozefowicz, B.~McGrew,
  J.~Pachocki, A.~Petron, M.~Plappert, G.~Powell, A.~Ray \emph{et~al.},
  ``Learning dexterous in-hand manipulation,'' \emph{The International Journal
  of Robotics Research}, vol.~39, no.~1, pp. 3--20, 2020.

\bibitem{bircher2021complex}
W.~G. Bircher, A.~S. Morgan, and A.~M. Dollar, ``Complex manipulation with a
  simple robotic hand through contact breaking and caging,'' \emph{Science
  Robotics}, vol.~6, no.~54, p. eabd2666, 2021.

\bibitem{hang2021manipulation}
K.~Hang, W.~G. Bircher, A.~S. Morgan, and A.~M. Dollar, ``Manipulation for
  self-identification, and self-identification for better manipulation,''
  \emph{Science Robotics}, vol.~6, no.~54, p. eabe1321, 2021.

\bibitem{morgan2022complex}
A.~Morgan, K.~Hang, B.~Wen, K.~E. Bekris, and A.~Dollar, ``Complex in-hand
  manipulation via compliance-enabled finger gaiting and multi-modal
  planning,'' \emph{IEEE Robotics and Automation Letters}, 2022.

\bibitem{higo2018rubik}
R.~Higo, Y.~Yamakawa, T.~Senoo, and M.~Ishikawa, ``Rubik's cube handling using
  a high-speed multi-fingered hand and a high-speed vision system,'' in
  \emph{2018 IEEE/RSJ International Conference on Intelligent Robots and
  Systems (IROS)}.\hskip 1em plus 0.5em minus 0.4em\relax IEEE, 2018, pp.
  6609--6614.

\bibitem{maekawa1995tactile}
H.~Maekawa, K.~Tanie, and K.~Komoriya, ``Tactile sensor based manipulation of
  an unknown object by a multifingered hand with rolling contact,'' in
  \emph{Proceedings of 1995 IEEE International Conference on Robotics and
  Automation}, vol.~1.\hskip 1em plus 0.5em minus 0.4em\relax IEEE, 1995, pp.
  743--750.

\bibitem{kumar2016learning}
V.~Kumar, A.~Gupta, E.~Todorov, and S.~Levine, ``Learning dexterous
  manipulation policies from experience and imitation,'' \emph{arXiv preprint
  arXiv:1611.05095}, 2016.

\bibitem{van2015learning}
H.~Van~Hoof, T.~Hermans, G.~Neumann, and J.~Peters, ``Learning robot in-hand
  manipulation with tactile features,'' in \emph{2015 IEEE-RAS 15th
  International Conference on Humanoid Robots (Humanoids)}.\hskip 1em plus
  0.5em minus 0.4em\relax IEEE, 2015, pp. 121--127.

\bibitem{falco2018policy}
P.~Falco, A.~Attawia, M.~Saveriano, and D.~Lee, ``On policy learning robust to
  irreversible events: An application to robotic in-hand manipulation,''
  \emph{IEEE Robotics and Automation Letters}, vol.~3, no.~3, pp. 1482--1489,
  2018.

\bibitem{liang2020hand}
J.~Liang, A.~Handa, K.~Van~Wyk, V.~Makoviychuk, O.~Kroemer, and D.~Fox,
  ``In-hand object pose tracking via contact feedback and gpu-accelerated
  robotic simulation,'' in \emph{2020 IEEE International Conference on Robotics
  and Automation (ICRA)}.\hskip 1em plus 0.5em minus 0.4em\relax IEEE, 2020.

\bibitem{koval2015pose}
M.~Koval, N.~Pollard, and S.~Srinivasa, ``Pose estimation for planar contact
  manipulation with manifold particle filters,'' \emph{The International
  Journal of Robotics Research}, vol.~34, no.~7, pp. 922--945, 2015.

\bibitem{zhang2012application}
L.~Zhang and J.~C. Trinkle, ``The application of particle filtering to grasping
  acquisition with visual occlusion and tactile sensing,'' in \emph{2012
  International Conference on Robotics and Automation}.\hskip 1em plus 0.5em
  minus 0.4em\relax IEEE, 2012.

\bibitem{bauza2020tactile}
M.~Bauza, E.~Valls, B.~Lim, T.~Sechopoulos, and A.~Rodriguez, ``Tactile object
  pose estimation from the first touch with geometric contact rendering,''
  \emph{arXiv preprint arXiv:2012.05205}, 2020.

\bibitem{navarro2021proximity}
S.~E. Navarro, S.~M{\"u}hlbacher-Karrer, H.~Alagi, H.~Zangl, K.~Koyama,
  B.~Hein, C.~Duriez, and J.~R. Smith, ``Proximity perception in human-centered
  robotics: A survey on sensing systems and applications,'' \emph{IEEE
  Transactions on Robotics}, 2021.

\bibitem{jiang2012seashell}
L.-T. Jiang and J.~R. Smith, ``Seashell effect pretouch sensing for robotic
  grasping,'' in \emph{2012 IEEE International Conference on Robotics and
  Automation}.\hskip 1em plus 0.5em minus 0.4em\relax IEEE, 2012, pp.
  2851--2858.

\bibitem{fang2019toward}
C.~Fang, D.~Wang, D.~Song, and J.~Zou, ``Toward fingertip non-contact material
  recognition and near-distance ranging for robotic grasping,'' in \emph{2019
  International Conference on Robotics and Automation (ICRA)}.\hskip 1em plus
  0.5em minus 0.4em\relax IEEE, 2019, pp. 4967--4974.

\bibitem{goger2013tactile}
D.~G{\"o}ger, H.~Alagi, and H.~W{\"o}rn, ``Tactile proximity sensors for
  robotic applications,'' in \emph{2013 IEEE International Conference on
  Industrial Technology (ICIT)}.\hskip 1em plus 0.5em minus 0.4em\relax IEEE,
  2013, pp. 978--983.

\bibitem{wistort2008electric}
R.~Wistort and J.~R. Smith, ``Electric field servoing for robotic
  manipulation,'' in \emph{2008 IEEE/RSJ International Conference on
  Intelligent Robots and Systems}.\hskip 1em plus 0.5em minus 0.4em\relax IEEE,
  2008, pp. 494--499.

\bibitem{mayton2010electric}
B.~Mayton, L.~LeGrand, and J.~R. Smith, ``An electric field pretouch system for
  grasping and co-manipulation,'' in \emph{2010 IEEE International Conference
  on Robotics and Automation}.\hskip 1em plus 0.5em minus 0.4em\relax IEEE,
  2010, pp. 831--838.

\bibitem{faller2019design}
L.-M. Faller, C.~Stetco, and H.~Zangl, ``Design of a novel gripper system with
  3d-and inkjet-printed multimodal sensors for automated grasping of a forestry
  robot,'' in \emph{2019 IEEE/RSJ International Conference on Intelligent
  Robots and Systems (IROS)}.\hskip 1em plus 0.5em minus 0.4em\relax IEEE,
  2019, pp. 5620--5627.

\bibitem{muhlbacher2015responsive}
S.~M{\"u}hlbacher-Karrer, A.~Gaschler, and H.~Zangl, ``Responsive
  fingers—capacitive sensing during object manipulation,'' in \emph{2015
  IEEE/RSJ International Conference on Intelligent Robots and Systems
  (IROS)}.\hskip 1em plus 0.5em minus 0.4em\relax IEEE, 2015, pp. 4394--4401.

\bibitem{guo2015transmissive}
D.~Guo, P.~Lancaster, L.-T. Jiang, F.~Sun, and J.~R. Smith, ``Transmissive
  optical pretouch sensing for robotic grasping,'' in \emph{2015 IEEE/RSJ
  International Conference on Intelligent Robots and Systems (IROS)}.\hskip 1em
  plus 0.5em minus 0.4em\relax IEEE, 2015, pp. 5891--5897.

\bibitem{hsiao2009reactive}
K.~Hsiao, P.~Nangeroni, M.~Huber, A.~Saxena, and A.~Y. Ng, ``Reactive grasping
  using optical proximity sensors,'' in \emph{2009 IEEE International
  Conference on Robotics and Automation}.\hskip 1em plus 0.5em minus
  0.4em\relax IEEE, 2009, pp. 2098--2105.

\bibitem{maldonado2012improving}
A.~Maldonado, H.~Alvarez, and M.~Beetz, ``Improving robot manipulation through
  fingertip perception,'' in \emph{2012 IEEE/RSJ International Conference on
  Intelligent Robots and Systems}.\hskip 1em plus 0.5em minus 0.4em\relax IEEE,
  2012, pp. 2947--2954.

\bibitem{yang2017pre}
B.~Yang, P.~Lancaster, and J.~R. Smith, ``Pre-touch sensing for sequential
  manipulation,'' in \emph{2017 IEEE International Conference on Robotics and
  Automation (ICRA)}.\hskip 1em plus 0.5em minus 0.4em\relax IEEE, 2017, pp.
  5088--5095.

\bibitem{patel2016integrated}
R.~Patel and N.~Correll, ``Integrated force and distance sensing using
  elastomer-embedded commodity proximity sensors.'' in \emph{Robotics: Science
  and systems}, 2016.

\bibitem{lancaster2019improved}
P.~E. Lancaster, J.~R. Smith, and S.~S. Srinivasa, ``Improved proximity,
  contact, and force sensing via optimization of elastomer-air interface
  geometry,'' in \emph{2019 International Conference on Robotics and Automation
  (ICRA)}.\hskip 1em plus 0.5em minus 0.4em\relax IEEE, 2019, pp. 3797--3803.

\bibitem{yamaguchi2017implementing}
A.~Yamaguchi and C.~G. Atkeson, ``Implementing tactile behaviors using
  fingervision,'' in \emph{2017 IEEE-RAS 17th International Conference on
  Humanoid Robotics (Humanoids)}.\hskip 1em plus 0.5em minus 0.4em\relax IEEE,
  2017, pp. 241--248.

\bibitem{sasaki2018robotic}
K.~Sasaki, K.~Koyama, A.~Ming, M.~Shimojo, R.~Plateaux, and J.-Y. Choley,
  ``Robotic grasping using proximity sensors for detecting both target object
  and support surface,'' in \emph{2018 IEEE/RSJ International Conference on
  Intelligent Robots and Systems (IROS)}.\hskip 1em plus 0.5em minus
  0.4em\relax IEEE, 2018, pp. 2925--2932.

\bibitem{konstantinova2016fingertip}
J.~Konstantinova, A.~Stilli, A.~Faragasso, and K.~Althoefer, ``Fingertip
  proximity sensor with realtime visual-based calibration,'' in \emph{2016
  IEEE/RSJ International Conference on Intelligent Robots and Systems
  (IROS)}.\hskip 1em plus 0.5em minus 0.4em\relax IEEE, 2016, pp. 170--175.

\bibitem{koyama2019high}
K.~Koyama, K.~Murakami, T.~Senoo, M.~Shimojo, and M.~Ishikawa, ``High-speed,
  small-deformation catching of soft objects based on active vision and
  proximity sensing,'' \emph{IEEE Robotics and Automation letters}, vol.~4,
  no.~2, pp. 578--585, 2019.

\bibitem{lancaster2022braking}
P.~Lancaster, C.~Mavrogiannis, S.~S. Srinivasa, and J.~R. Smith,
  ``Electrostatic brakes enable individual joint control of underactuated,
  highly articulated robots,'' \emph{In Submission}.

\bibitem{thrun2001robust}
S.~Thrun, D.~Fox, W.~Burgard, and F.~Dellaert, ``Robust monte carlo
  localization for mobile robots,'' \emph{Artificial intelligence}, vol. 128,
  no. 1-2, pp. 99--141, 2001.

\bibitem{makoviychuk2021isaac}
V.~Makoviychuk, L.~Wawrzyniak, Y.~Guo, M.~Lu, K.~Storey, M.~Macklin,
  D.~Hoeller, N.~Rudin, A.~Allshire, A.~Handa \emph{et~al.}, ``Isaac gym: High
  performance gpu-based physics simulation for robot learning,'' \emph{arXiv
  preprint arXiv:2108.10470}, 2021.

\bibitem{williams2017information}
G.~Williams, N.~Wagener, B.~Goldfain, P.~Drews, J.~M. Rehg, B.~Boots, and E.~A.
  Theodorou, ``Information theoretic mpc for model-based reinforcement
  learning,'' in \emph{2017 IEEE International Conference on Robotics and
  Automation (ICRA)}.\hskip 1em plus 0.5em minus 0.4em\relax IEEE, 2017, pp.
  1714--1721.

\bibitem{zhong2019}
\BIBentryALTinterwordspacing
J.~Zhong, M.~Oller, T.~Power, and K.~Taekyung, ``{PyTorch MPPI}.'' [Online].
  Available: \url{https://github.com/UM-ARM-Lab/pytorch_mppi}
\BIBentrySTDinterwordspacing

\end{thebibliography}
\bibliographystyle{IEEEtran}

\end{document}